\documentclass[preprint,12pt,authoryear]{elsarticle}

\usepackage{hyperref}
\hypersetup{
    colorlinks=true,
    linkcolor=blue,
    filecolor=magenta,
    urlcolor=cyan,
    citecolor=blue,
}
\usepackage{multirow}
\usepackage{booktabs}
\usepackage{caption}

\usepackage{amssymb}

\usepackage{lineno}

\begin{document}

\begin{frontmatter}



\title{HrSegNet : Real-time High-Resolution Neural Network with Semantic Guidance for Crack Segmentation \tnoteref{lable1}}




\author[label1]{Yongshang Li \corref{cor1} }
\ead{yshli@chd.edu.cn}
\cortext[cor1]{Corresponding authors}

\author[label1]{Ronggui Ma \corref{cor1}}
\ead{rgma@chd.edu.cn}

\author[label1]{Han Liu}

\author[label2]{Gaoli Cheng}

\affiliation[label1]{organization={School of Information Engineering, Chang'an University},
            city={Xi'an},
            postcode={710064}, 
            state={Shaanxi},
            country={China}}
            
\affiliation[label2]{organization={Shaanxi Expressway Mechanisation Engineering Co.,Ltd},
            city={Xi'an},
            postcode={710038}, 
            state={Shaanxi},
            country={China}}

\begin{abstract}
Through extensive research on deep learning in recent years and its application in construction, crack detection has evolved rapidly from rough detection at the image-level and patch-level to fine-grained detection at the pixel-level, which better suits the nature of this field. Despite numerous existing studies utilizing off-the-shelf deep learning models or enhancing them, these models are not always effective or efficient in real-world applications. In order to bridge this gap, we propose a High-resolution model with Semantic guidance, specifically designed for real-time crack segmentation, referred to as HrSegNet.
Our model maintains high resolution throughout the entire process, as opposed to recovering from low-resolution features to high-resolution ones, thereby maximizing the preservation of crack details. Moreover, to enhance the context information, we use low-resolution semantic features to guide the reconstruction of high-resolution features. To ensure the efficiency of the algorithm, we design a simple yet effective method to control the computation cost of the entire model by controlling the capacity of high-resolution channels, while providing the model with extremely strong scalability.
Extensive quantitative and qualitative evaluations demonstrate that our proposed HrSegNet has exceptional crack segmentation capabilities, and that maintaining high resolution and semantic guidance are crucial to the final prediction. Compared to state-of-the-art segmentation models, HrSegNet achieves the best trade-off between efficiency and effectiveness. Specifically, on the crack dataset CrackSeg9k, our fastest model HrSegNet-B16 achieves a speed of 182 FPS with 78.43\% mIoU, while our most accurate model HrSegNet-B48 achieves 80.32\% mIoU with an inference speed of 140.3 FPS. Furthermore, the quantitative results demonstrate that our model maintains robustness and stability in the presence of noisy data.

\end{abstract}

\begin{graphicalabstract}
\includegraphics[scale=0.4]{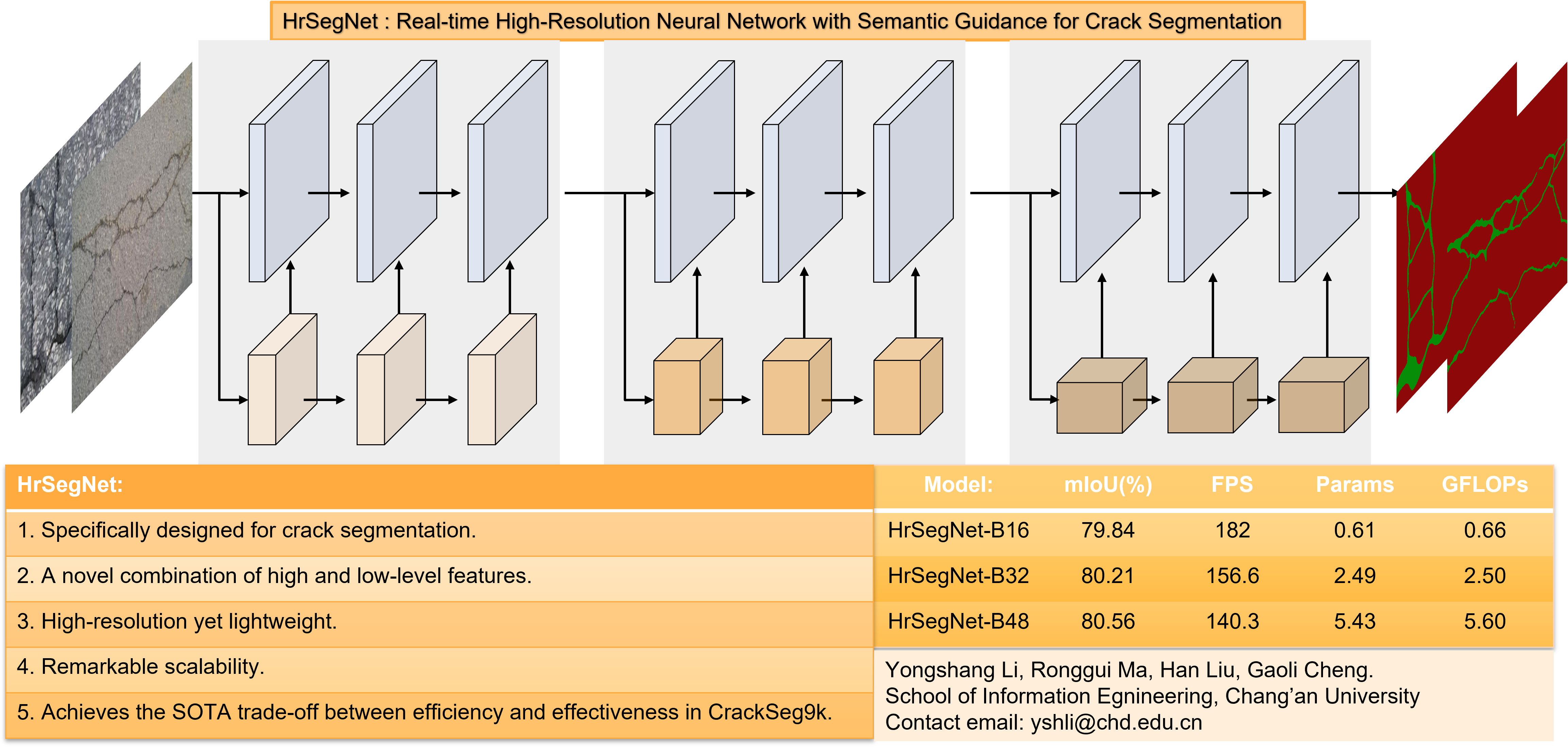}
\end{graphicalabstract}

\begin{highlights}
\item HrSegNet is a high-resolution model explicitly designed for real-time crack segmentation, maintaining high resolution throughout the process to maximize the preservation of crack details.
\item The proposed model uses low-resolution semantic features to guide the reconstruction of high-resolution features, enhancing the context information in the model and improving the final segmentation.
\item The architecture includes a simple yet efficient method to control the entire model's computation cost by controlling the high-resolution channel's capacity, providing strong scalability while maintaining efficiency.
\item HrSegNet achieves the best trade-off between efficiency and effectiveness compared to current popular segmentation models. The fastest model, HrSegNet-B16, achieves an inference speed of 182 FPS and 78.43\% mIoU on the benchmark crackSeg9k, with a computational complexity of 0.66 GFLOPs. The model with the highest accuracy, HrSegNet-B48, achieves 80.32\% mIoU at 140.3 FPS, with a computational complexity of 5.60 GFLOPs.
\end{highlights}


\end{frontmatter}


\section{Introduction}
\label{sec:intro}









Cracks are early ailments of buildings, bridges, and highways \citep{hsieh_machine_2020}. Timely detection and repair can mitigate subsequent maintenance costs and ensure the user's safety. Traditional methods for crack detection, such as visual inspection and manual assessment, are costly, inefficient, and susceptible to subjective errors resulting in missed or false detections. Non-contact detection techniques evaluate cracks or defects in the target without physical contact \citep{ryuzono2022performance, wu2023learning, xia2023eddy}. These methods surpass manual approaches in precision and efficiency but heavily rely on equipment and require specialized knowledge. The advancement of digital image processing techniques has significantly expedited crack detection; however, the results are influenced by image quality, including noise that diminishes detection accuracy. Furthermore, the robustness of digital image processing techniques is weak when facing challenges posed by complex environments characterized by low lighting, reflections, and deformations \citep{munawar_image-based_2021}.

The advent of deep learning methods, particularly convolutional neural networks (CNNs), heralds a breakthrough in image processing techniques. Due to the efficiency, accuracy, and end-to-end capabilities, an increasing number of researchers are applying CNNs to the field of crack detection. CNN-based crack detection methods can be classified into three categories: image-level classification, patch-level object detection, and pixel-level semantic segmentation \citep{hsieh_machine_2020}. The first two methods can locate the position of cracks in an image, but their results are coarse and cannot determine the morphology and quantification of the cracks. Semantic segmentation assigns a label to each pixel in the image, enabling precise localization of crack pixels. As a result, it is naturally suited for crack detection tasks.


Most existing crack segmentation methods adopt models based on general scene understanding, overlooking the challenges specific to crack segmentation tasks in practical applications. Crack segmentation tasks differ from general scene-agnostic segmentation tasks. In general scene images, such as COCO-stuff \citep{caesar_coco-stuff_2018} and Cityscapes \citep{cordts_cityscapes_2016}, multiple object classes of interest have similar pixel proportions. However, in crack images, the proportion of pixels representing the objects of interest is merely 1\% of all pixels \citep{xu_pixel-level_2021}. This gives rise to a highly imbalanced pixel-level classification task. Furthermore, cracks can exhibit diverse shapes, occur in complex backgrounds, and frequently coexist with noise, further complicating the task.

\begin{figure}[t]
    \centering
    \includegraphics[scale=0.6]{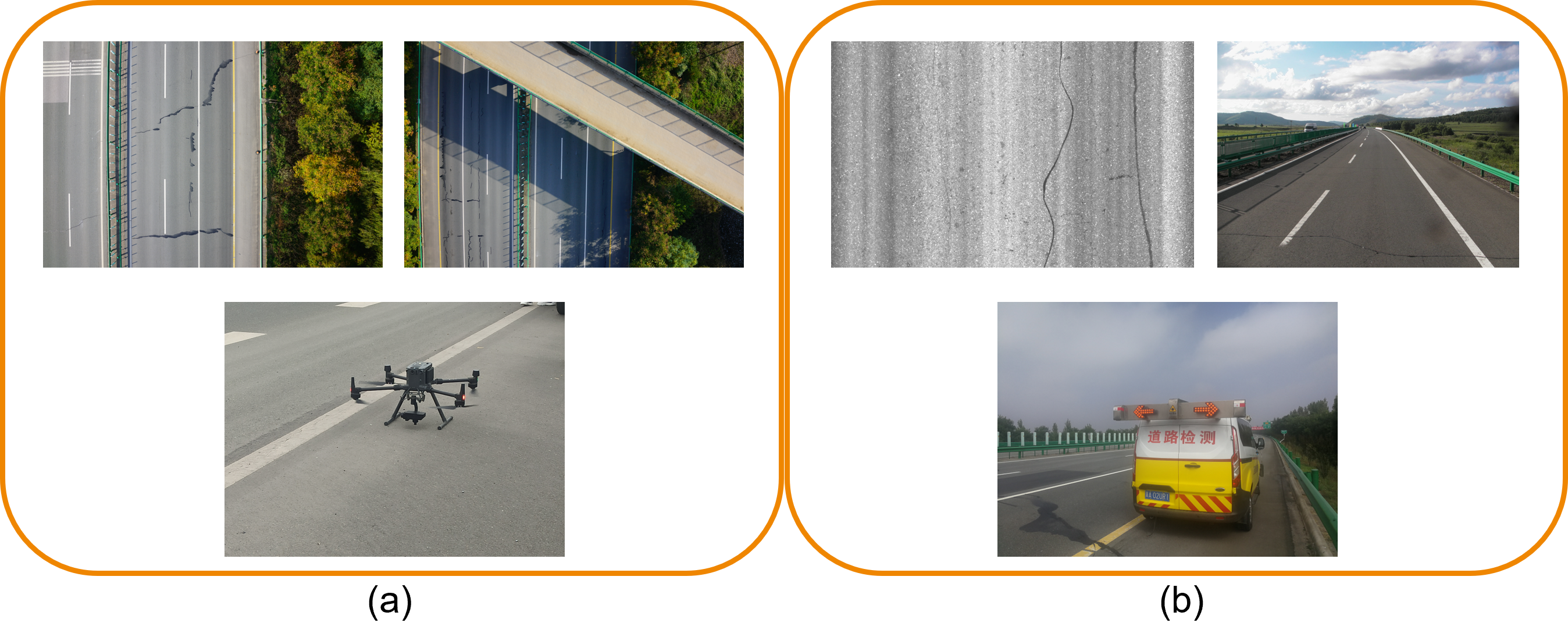}
    \caption{Automated inspection apparatuses and their data: (a) unmanned aerial vehicle; (b) road measurement vehicle.}
    \label{fig:fig0}
\end{figure}

Current crack detection tasks increasingly rely on fast detection devices such as drones \citep{ding_crack_2023}, road measurement vehicles \citep{guo_pavement_2023}, and specially customized robots \citep{kouzehgar2019self}, as shown in Figure \ref{fig:fig0}. These edge devices prioritize lightweight and real-time processing, often lacking high computational power. Therefore, there are strict requirements for algorithm complexity and efficiency. Several studies have found that high-resolution CNNs possess a superior ability to capture fine details and perform well in location-sensitive tasks \citep{wang_deep_2020, xu_pixel-level_2021, wang_u-hrnet_2022, jia_efficient_2022, zhang_recurrent_2022}. However, high-resolution features significantly increase computational cost and model complexity, making it challenging for such models to meet real-time demands in practical crack segmentation. Based on these observations, we identify a gap between current CNN-based crack segmentation models and the real-time application in the real-world.

We propose a real-time high-resolution model, HrSegNet, to achieve high performance and efficiency in crack segmentation. Our model includes a high-resolution path designed to extract detailed information while maintaining high resolution throughout, as well as an auxiliary semantic path that provides step-by-step contextual guidance and enhancement to the high-resolution path. To ensure real-time performance while controlling computational cost, we control the channel capacity of the entire high-resolution path, thereby making the model highly lightweight and scalable. HrSegNet uses a two-stage segmentation head to restore resolution incrementally rather than in one step, thereby improving segmentation accuracy at a small computational cost. HrSegNet achieves superior accuracy while maintaining real-time performance, as evidenced by extensive experimental results on two crack benchmarks \citep{benz_crack_2019, kulkarni_crackseg9k_2022}.

The main contributions can be summarized as follows:
\begin{itemize}
    \item A high-resolution model explicitly designed for crack segmentation, which enhances detailed features with semantic guidance while maintaining high resolution throughout the process.
    \item We design the HrSegNet to be highly scalable, enabling a lightweight backbone for a breakneck inference speed or increased channel capacity for improved accuracy.
    \item The fastest model we proposed, HrSegNet-B16, achieves an inference speed of 182 FPS and 78.43\% mIoU on the benchmark CrackSeg9k, with a computational complexity of 0.66 GFLOPs. The model with the highest accuracy, HrSegNet-B48, achieves 80.32\% mIoU at 140.3 FPS, with a computational complexity of 5.60 GFLOPs.
    \item The code, trained weights, and training records of the models are publicly available at \href{https://github.com/CHDyshli/HrSegNet4CrackSegmentation}{https://github.com/CHDyshli/HrSegNet4CrackSegme\\ntation}
\end{itemize}

The rest of this paper is organized as follows. Section \ref{sec:relatedwork} presents the current studies relevant to this study. The methodology are described in Section \ref{sec:method}. The experiments and results are outlined in Section \ref{sec:Experimentsandresults}. Lastly, we summarize all of the work.


\section{Related work}
\label{sec:relatedwork}

Deep learning-based semantic segmentation has dramatically advanced the performance of crack detection. The cutting-edge research mainly explores three directions: higher segmentation accuracy, faster inference speed, and more effective feature fusion. Therefore, this section will introduce crack segmentation-related work from these three aspects.

\subsection{High-resolution models}
\label{subsec:highper}


Many studies indicate that high-resolution representation is essential for detecting small objects, such as cracks \citep{chen_automatic_2021, xu_pixel-level_2021, jia_efficient_2022, zhang_recurrent_2022}. HRNet \citep{wang_deep_2020} adopted a high-resolution design, decomposing the feature extraction and fusion processes into different branches, which maintains high-resolution and multi-scale features. \citet{xu_pixel-level_2021} and \citet{zhang_recurrent_2022} aimed to deal with high-resolution crack images and strove to maintain the integrity of details, then they used HRNet as the baseline model. \citet{tang_semantic_2021} proposed using higher-resolution feature maps to solve the grid effect problem caused by dilated convolution in deep neural networks. 
Given the heavy nature of the original HRNet backbone, \citet{chen_automatic_2021} opted to eliminate the down-sampling layer in the initial stage while reducing the number of high-resolution representation layers. Furthermore, integrating dilated convolution and hierarchical features were introduced to decrease the model's parameters while maintaining accuracy.
\citet{xiao_pavement_2023} innovatively proposed a high-resolution network structure based on the transformer to more reasonably utilize and fuse multi-scale semantic features.

Although the abovementioned approaches can achieve high accuracy, they come at the cost of high computational consumption and latency. This is because high-resolution feature maps result in more convolutional operations, which dominate the model's complexity. To achieve real-time performance, models require low-latency inference, which is not feasible with high-precision ones.


\subsection{Real-time models}
\label{subsec:realtime}

Most methods use lightweight backbone to achieve real-time crack segmentation. A lightweight encoder-decoder model called LinkCrack was designed based on the UNet \citep{liao_automatic_2022}. The authors adopted a ResNet34 with reduced channel numbers for the encoder, resulting in an inference speed of 17 FPS and 3.4 M parameters. 
\citet{jiang_two-step_2022} proposed an improved DeeplabV3+ for road crack segmentation. The authors modified the encoder of the original architecture and introduced Ghost modules from GhostNet to generate more Ghost feature maps. This reduced the parameters required for forward propagation and computational complexity while maintaining performance. \citet{yong_riianet_2022} proposed a novel approach to address the inefficiency of current mainstream CNNs, which overlooks the importance of different-level feature extractors. They introduced an asymmetric convolution enhancement module for low-level feature extraction and a residual expanded involution module for high-level semantic enhancement in crack segmentation task.


\subsection{Feature fusion}
\label{subsec:feature fusion}

In the context of semantic segmentation models, it is commonly agreed that the fusion of features from different scales is crucial for achieving accurate results. Currently, two main approaches for feature fusion based on their location are cross-layer connections and pyramid pooling. The typical model for cross-layer connections is UNet \citep{ronneberger_u-net_2015}, which extracts features from different layers through a completely symmetric encoder-decoder structure.
\citet{huyan_pixelwise_2022} compared two models, VGG-UNet and Res-UNet, which utilize VGG and ResNet as backbones respectively.  
\citet{xu_pavement_2022-2} designed an encoder-decoder model similar to UNet for crack segmentation in CCD images. They introduced the transformer to capture long-range contextual features in the image instead of using convolution. 
Pyramid pooling \citep{liu_deepcrack_2019} and atrous spatial pyramid pooling \citep{sun_dma-net_2022, xu_pixel-level_2021, tang_semantic_2021} are used to model long-range contextual information and extract features of different scales. 

High-resolution detailed features are crucial for crack segmentation, but contextual information can still assist the model in achieving more accurate segmentation. Therefore, we propose a fusion method called ``semantic guidance" that compensates for detailed information with semantic information, as discussed in the Section \ref{subsec:Semanticguidance}. Our method differs entirely from cross-layer connections and pyramid pooling because we extract low-level features and fuse high-level information simultaneously. This parallel processing approach makes our model more efficient.

\section{Method}
\label{sec:method}

The concept behind the proposed model is intuitive, with particular emphasis on crack detection. Our design philosophy is based on three key points: (1) high-resolution representations are crucial for detecting small objects such as cracks; (2) semantic features can guide and strengthen the extraction of comprehensive contextual information from high-resolution representations; (3) high-resolution means high computational costs, so it is necessary to control that in order to achieve real-time segmentation. 

\begin{figure}[t]
    \centering
    \includegraphics[width=\textwidth]{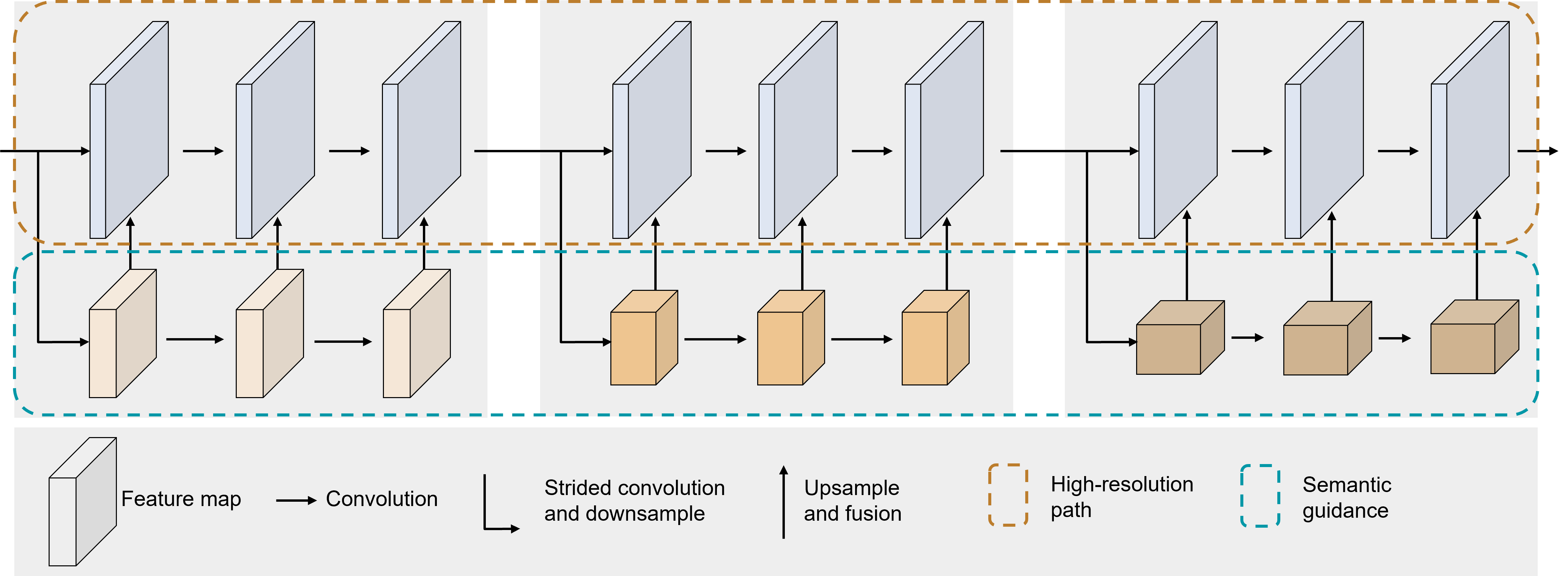}
    \caption{The main body of the proposed HrSegNet.}
    \label{fig:fig1}
\end{figure}

\subsection{High-resolution path}
\label{subsec:Highresolutionpath}
In tasks requiring attention to detail and location sensitivity, high-resolution representation is of paramount importance. Nevertheless, high resolution entails a concurrent increase in computational demand.

Inspired by the ideas from STDCNet \citep{fan_rethinking_2021} and HRNet \citep{wang_deep_2020}, we design a simple, efficient, and controllable high-resolution path to encode rich detail information in crack images. As shown in Figure \ref{fig:fig1}, the high-resolution path contains three High-resolution with Semantic Guidance (HrSeg) blocks and maintains the identical resolution throughout the process. However, ordinary convolutions are very expensive when faced with high-resolution feature maps. When convolution is applied to high spatial resolution, the floating-point operations (FLOPs) are dominated by the spatial size of the output feature map. For ordinary convolution, given input and output channel numbers, $C_{in}$ and $C_{out}$, kernel size $k$, and output's spatial size $W_{out}*H_{out}$, when ignoring bias, the FLOPs of the convolution can be represented as:

\begin{equation}
    FLOPs = C_{in} * C_{out} * k * k * W_{out} * H_{out} \label{eq:conv_flops}
\end{equation}

In our design, the convolutional kernel size $k$ and the output feature size $W_{out}*H_{out}$ remain constant. Therefore, we can control the FLOPs by defining $C_{in}$ and $C_{out}$. In our setting, we set $C_{in}$ equal to  $C_{out}$, and the default value is not greater than 64. This effectively controls the computational cost.

As shown in Figure \ref{fig:fig1}, our high-resolution path consists of three stages, each containing three layers. Each layer includes a convolution with stride 1, followed by Batch Normalization (BN) and ReLU. It should be noted that we omit the stem of the model in Figure \ref{fig:fig1}. The stem consists of two Conv-BN-ReLU sequences, each of which down-sample the spatial resolution of the input by a factor of 2. Therefore, before entering the high-resolution path, the size of the feature map is 1/4 of the original image, and the channel number and spatial resolution remain unchanged throughout the subsequent process.

\subsection{Semantic guidance}
\label{subsec:Semanticguidance}

It is commonly believed that high-resolution feature maps contain rich details while down-sampling provides a sufficient receptive field for extracting contextual semantic information.
A dedicated context path is used to obtain macro features in the two-stream model, BiSeNetV2 \citep{yu_bisenet_2020}. However, the dual path causes information and structural redundancy, leading to inefficiency. HRNet \citep{wang_deep_2020} designs a gradually increasing sub-network from high to low resolution and parallel connects multi-resolution branches. However, its model is too heavy and unwieldy and far exceeds the requirements of real-time inference.

\begin{figure}[t]
    \centering
    \includegraphics[width=\textwidth]{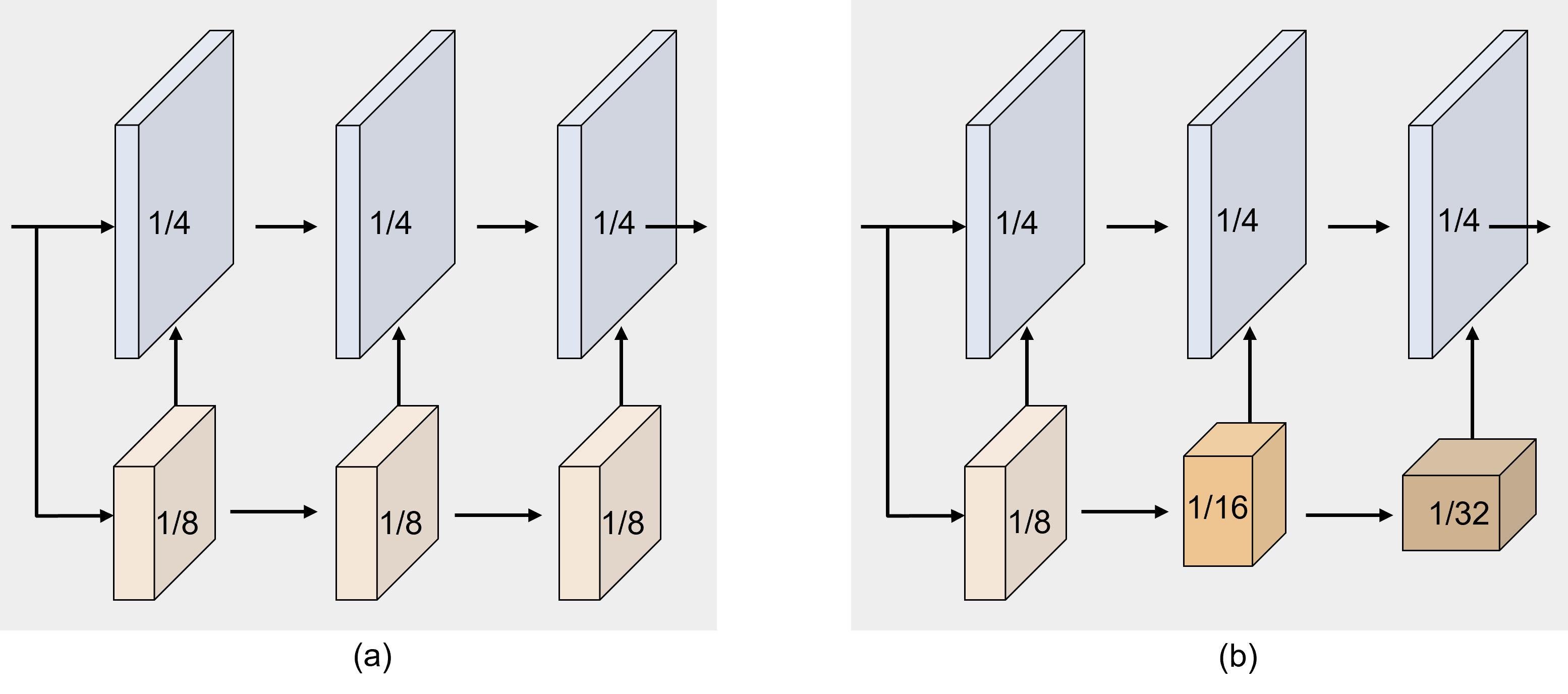}
    \caption{Two examples of the HrSeg block. (a) illustrates the semantic-guided component within the HrSeg block, which maintains the same resolution as the high-resolution path but gradually decreases by a factor of 2 in subsequent blocks. (b) demonstrates another way to provide semantic guidance by gradually decreasing the spatial resolution of the semantic guidance within a HrSeg block. }
    \label{fig:fig2}
\end{figure}

To address the issue of redundancy caused by separate context paths, as seen in BiSeNetV2 and HRNet, we propose a parallel semantic guidance path that is lightweight and flexible. Our approach involves down-sampling the high-resolution features and fusing them with semantic features for guidance and assistance simultaneously throughout the feature reconstruction process. The HrSeg block we designed, shown in Figure \ref{fig:fig2}, demonstrates this process. Our design allows for flexible adjustment of semantic guidance, such as using different (Figure \ref{fig:fig2} (a)) or identical (Figure \ref{fig:fig2} (b)) down-sampling manners in the same block or different fusion methods during feature aggregation.
Figure \ref{fig:fig2} (a) illustrates the semantic-guided component within the HrSeg block, which maintains the identical resolution as the high-resolution path but gradually decreases by a factor of 2 in subsequent blocks.
Figure \ref{fig:fig2} (b) demonstrates another way to provide semantic guidance by gradually decreasing the spatial resolution of the semantic guidance within a HrSeg block. Each semantic-guided feature map is up-sampled to the same size as the high-resolution path and then adjusted to the same number of channels via a 1 × 1 convolution. The different down-sampling and fusion strategies will be discussed in Section \ref{subsubsec:semanticguidance} and \ref{subsubsec:featurefusion}.

\subsection{Segmentation head}
\label{subsec:Segmentationhead}

Many semantic segmentation models with encoder-decoder structures usually perform aggregation of features at different levels before the final segmentation \citep{howard_searching_2019, peng_pp-liteseg_2022}. However, since we have continuously fused features at intermediate layers while maintaining high resolution throughout, the output directly enters the segmentation head.

\begin{figure}[t]
    \hspace{-2cm}
    \centering
    \includegraphics[scale=0.5]{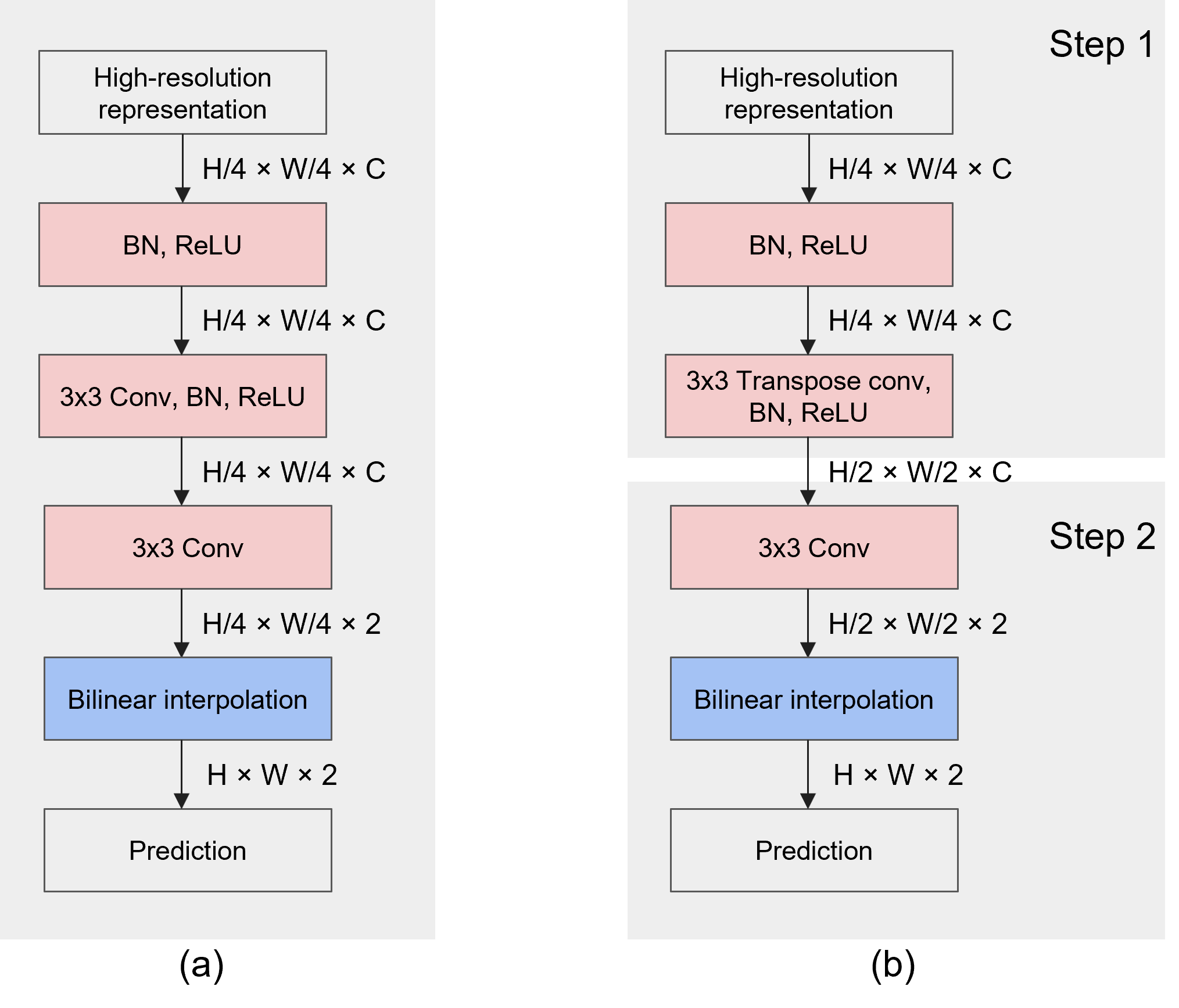}
    \caption{(a) is the single-step segmentation head. (b) is the two-step segmentation head.}
    \label{fig:fig3}
\end{figure}

We gradually recover the original spatial resolution from the high-resolution representation in steps instead of directly restoring from a 1/8-sized feature map to the original image size, as many existing works do (see Figure \ref{fig:fig3} (a)). Our approach, as shown in Figure \ref{fig:fig3} (b), first applies a 3 × 3 transposed convolution to the high-resolution representation, restoring spatial resolution to half the size of the original image. In the second step, the previous features are restored to the original image size through bilinear interpolation. The comparison between the single-step and double-step manners is illustrated in Section \ref{subsubsec:segmentationhead}.

\subsection{Deep supervision}
\label{subsec:Deep supervision}

\begin{figure}[t]
    \centering
    \includegraphics[scale=0.5]{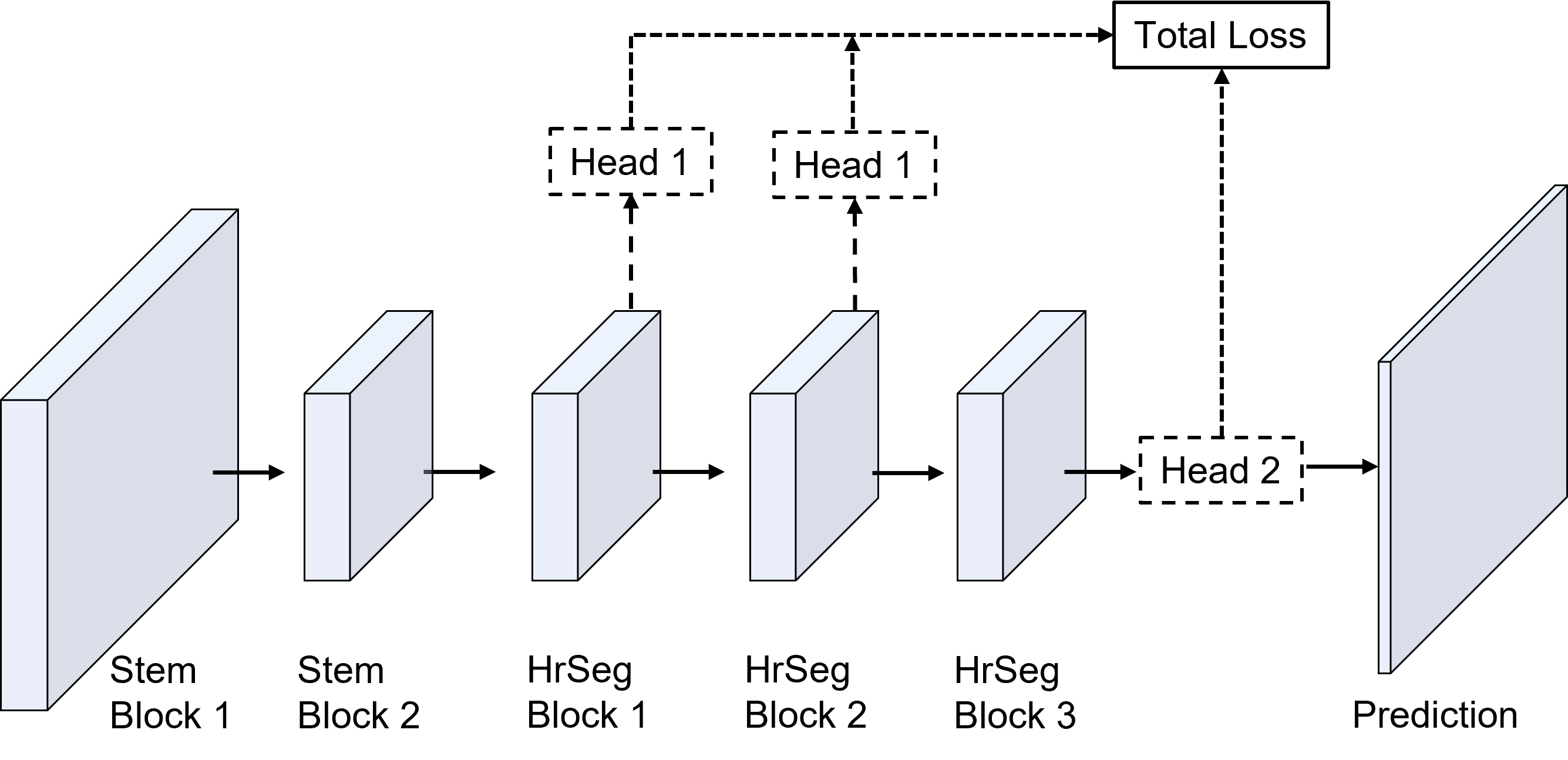}
    \caption{Deep supervision utilized in HrSegNet. Head 1 is a single-step segmentation head, whereas Head 2 is a double-step segmentation head.}
    \label{fig:fig4}
\end{figure}

Additional supervision can facilitate the optimization of deep CNNs during the training process. PSPNet \citep{zhao2017pyramid} demonstrates the effectiveness of this approach by adding auxiliary loss at the output of the res4\_22 block in ResNet-101 and setting the corresponding weights to 0.4. BiSeNetV2 \citep{yu_bisenet_2020} proposes booster training, which involves adding extra segmentation heads at the end of each stage in the semantic branch. 

We add auxiliary loss to the final convolution layer of each HrSeg block, as shown in Figure \ref{fig:fig4}. Unlike the final primary loss, the auxiliary loss segmentation heads follow the scheme shown in Figure \ref{fig:fig3} (a). During the inference stage, the auxiliary heads are ignored, thus not affecting the overall inference speed. The total loss is the weighted sum of the cross-entropy loss of each segmentation head, as shown in Equation (\ref{eq:eq2}):
\begin{equation}
    L_t = L_p + \alpha \sum_{i=1}^n L_i
    \label{eq:eq2}
\end{equation}
\(L_t\), \(L_p\), and \(L_i\) represent the total loss, primary loss, and auxiliary loss, respectively. In this work, the number of auxiliary loss $n$ is 2, and the weight $\alpha$ is set to 0.5.

\subsection{Overall architecture}
\label{subsec:Overallarchitecture}

Table \ref{tab:table1} presents an instance of HrSegNet. Each stage consists of a set of convolution operations, with each operation containing the parameters kernel size $k$, output channel $c$, and stride $s$. The default value of $c$ is set to base, which is a constant that controls the computational complexity.
\begin{center}
\begin{table}[t]
    \centering
    \begin{tabular}{ c|c c c c|c c c c|c } 
\toprule
\multirow{2}{*}{Stage}& \multicolumn{4}{c|}{High-resolution path} & \multicolumn{4}{c}{Semantic guidance }& Output size\\ 
\cline{2-10}
 & \textit{opr} & \textit{k} & \textit{c} & \textit{s} & \textit{opr} & \textit{k} & \textit{c} & \textit{s}  & \\ 
 \hline    
 Input & \multicolumn{4}{c|}{-} & \multicolumn{4}{c|}{-} & 400×400×3\\
 \hline
 Stem block 1 & Conv2d & 3×3 & base & 2 &  \multicolumn{4}{c|}{-} & 200×200×base\\
 \hline
 Stem block 2 & Conv2d & 3×3 & base & 2 &  \multicolumn{4}{c|}{-} & 100×100×base\\
 \hline
 \multirow{3}{*}{HrSeg block 1}  & Conv2d & 3×3 & base & 1 &  Conv2d & 3×3 & base×2 & 2 & \\

& Conv2d & 3×3 & base & 1 &  Conv2d & 3×3 & base×2 & 1 & 100×100×base\\

& Conv2d & 3×3 & base & 1 &  Conv2d & 3×3 & base×2 & 1 & \\
\hline
 \multirow{3}{*}{HrSeg block 2}  & Conv2d & 3×3 & base & 1 &  Conv2d & 3×3 & base×4 & 2 & \\

& Conv2d & 3×3 & base & 1 &  Conv2d & 3×3 & base×4 & 1 & 100×100×base\\

& Conv2d & 3×3 & base & 1 &  Conv2d & 3×3 & base×4 & 1 & \\
\hline
 \multirow{3}{*}{HrSeg block 3}  & Conv2d & 3×3 & base & 1 &  Conv2d & 3×3 & base×8 & 2 & \\

& Conv2d & 3×3 & base & 1 &  Conv2d & 3×3 & base×8 & 1 & 100×100×base\\

& Conv2d & 3×3 & base & 1 &  Conv2d & 3×3 & base×8 & 1 & \\
\hline
\multirow{3}{*}{Seg head} & Trans. Conv2d & 3×3 & base & 2 & \multicolumn{4}{c|}{} & 200×200×base\\
& Conv2d & 3×3 & 2 & 1 & \multicolumn{4}{c|}{-}& 400×400×2\\
& \multicolumn{4}{|c|}{Bilinear interpolation}& \multicolumn{4}{c|}{} & 400×400×2 \\
\bottomrule
\end{tabular}

    \caption{Instantiation of the HrSegNet. \textit{Opr} represents different operations. Each operation has a kernel size \textit{k}, stride \textit{s}, and output channel \textit{c}. Conv2d denotes a combination of Conv-BN-ReLU. Trans. Conv2d represents a transposed convolution. }
    
    \label{tab:table1}
\end{table}
\end{center}

The model comprises six stages, with each of the first two stages containing a stem block consisting of a Conv-BN-ReLU sequence with a stride of 2. The stem blocks quickly reduce the spatial dimensions of the input image to 1/4, with a feature map channel base. To reduce the computations of high-resolution representation, we assigned each stem block only one convolution, which has been proven to be sufficient in subsequent experiments. The second, third, and fourth stages are our carefully crafted HrSeg blocks. Each HrSeg block contains a high-resolution path and a semantic guidance branch. The feature map size of the high-resolution path remains unchanged throughout, while that of the semantic guidance path gradually decreases as the channel number increases. We use the same style as ResNet where channel numbers double when spatial resolution is halved. The final stage is the segmentation head, where the feature map from the previous layer is restored to the original size through a transposed convolution and bilinear interpolation. As we only predict cracks and background, the predicted output channel is 2.


In our experiments, we studied three models: HrSegNet-B16, HrSegNet-B32, and HrSegNet-B48, where 16, 32, and 48 represent the channel numbers of the high-resolution path. By managing the size of the base, we control the computational complexity of the model, making it highly scalable.

\section{Experiments and results}
\label{sec:Experimentsandresults}


This section will first introduce the datasets and evaluation metrics. Subsequently, we will provide a comprehensive depiction of the experimental setup. We scrutinize the significance and influence of each component in the HrSegNet and assess the scalability and generalization aptitude. Finally, we compare the accuracy and speed of HrSegNet with state-of-the-art.

\subsection{Datasets and evaluation metrics}
\label{subsec:datasets}

In the field of crack segmentation, publicly available datasets are relatively small compared to general scenarios in size and number, making it difficult to establish a fair benchmark for algorithm comparison. Currently, two works integrate previous crack datasets: CrackSeg9k \citep{kulkarni_crackseg9k_2022} and crack segmentation dataset \citep{benz_crack_2019}. They contain seven identical sub-datasets. However, there are still some differences. CrackSeg9k was refined to address the presence of noisy annotations, while the latter consists of raw images without preprocessing. For convenience, we refer to them as the Original Crack Dataset (OCD) and the Refined Crack Dataset (RCD) throughout the rest of this paper. 
OCD has 9,887 images (448 × 448 resolution), and RCD has 9,255 images (400 × 400 resolution). Both datasets have background and crack labels but lack designated training, validation, and test sets. To ensure a fair comparison, we randomly select 900 images each for validation and testing from shared images, with the rest used for training. OCD is noisier than RCD, so we use it to evaluate the model's generalization ability while we choose RCD to evaluate theoretical performance.

We employ two evaluation metrics to assess the segmentation performance, namely mean Intersection over Union (mIoU) used to assess accuracy and Frames Per Second (FPS) as a measure of speed. In addition, the floating-point operations (FLOPs) and parameters (Params) of the model serve as indicators to evaluate the computational complexity and size.



\subsection{Implementation details}
\label{subsec:implem}

The training phase on OCD and RCD datasets employs mini-batch stochastic gradient descent with a momentum of 0.9 and weight decay of 5e-4. The batch size is set to 32. A ``poly" policy is used to control the learning rate where the initial rate of 0.01 is multiplied by $ (1 - \frac{iter}{max\_iter})^{power}$, with the power set to 0.9. The models are trained for 100,000 iterations from scratch with ``kaiming normal" initialization. A warm-up strategy is used for the first 2000 iterations to ensure stable training. We use various data augmentation techniques, including random distortion, random horizontal flipping, random cropping, random resizing, and normalization. The scale range for random resizing is consistent between the two datasets, as both use a range of 0.5 to 2.0. The random distortion applies random variations to an image's brightness, contrast, and saturation levels, with each parameter set to 0.5. All the training images are cropped to 400 × 400 resolution. Online Hard Example Mining (OHEM) is used to train all models.

We run the models using TensorRT for a fair comparison during the inference phase. For the OCD, the data is resized to 400 × 400 for and then restored to original 448 × 448. The inference time is measured using an NVIDIA GeForce RTX 2070 SUPER with CUDA 12.1 and cuDNN 8.9. The inference process is carried out over ten iterations to reduce the impact of error fluctuations. We conduct all experiments based on Paddle 2.4 and the same hardware platform.

\subsection{Ablation study on RCD}
\label{subsec:ablation}

In this subsection, we conducted an ablation study on the RCD to evaluate the effectiveness of the components of HrSegNet.

\subsubsection{High-resolution path only}

We first explore the influence of resolution on crack segmentation results in the high-resolution path. HRNet \citep{wang_deep_2020} and DDRNet \citep{pan_deep_2023} keep the high-resolution branch at 1/4 and 1/8 of the original image resolution, respectively, in order to extract detailed features. Previous work has yet to attempt to maintain the high-resolution path at the general original image resolution, as the convolutional operations in the high-resolution path consume too much computation. However, as our high-resolution path, which controls computational cost by managing channel numbers, is very lightweight, we also attempt a high-resolution path with a 1/2 original image resolution. As discussed in Section \ref{subsec:Highresolutionpath}, we control the high-resolution path's spatial resolution by defining the stem's output. Table \ref{tab:table2} shows detailed comparative experiments of three different resolutions. When the resolution is set to 1/4 of the original image, the high-resolution model achieves 74.03\% mIoU, which is 3.6\% and 1.33\% higher than that of 1/2 and 1/8, respectively. Although the accuracy of 1/8 resolution is inferior to 1/4 at 1.33\%, the computation is only 28\% of the former. When the computational requirements of the running device are extremely stringent, 1/8 resolution is still an excellent choice. However, in subsequent experiments, we still choose the best accuracy of 1/4 resolution as our default.

\begin{center}
    \begin{table}[t]
        \centering
        \begin{tabular}{c c c| c c| c |c |c}
             \toprule
             \multicolumn{3}{c|}{HR path} & \multicolumn{2}{c|}{SG path} & \multirow{2}{*}{mIoU(\%)} & \multirow{2}{*}{GFLOPs} & \multirow{2}{*}{Params(M)}\\
             \cline{1-5}
             \textit{1/2} & \textit{1/4} & \textit{1/8} & \textit{single} & \textit{multi} & & \\
             \hline
             \checkmark & & & & & 70.43 & 5.06 & 0.099\\
              &\checkmark & & & & 74.03 & 1.28 & 0.099\\
              & &\checkmark & & & 72.70 & 0.36 & 0.101\\

              \hline
              & \checkmark & & \checkmark & & \textbf{76.75} & 2.31 & 2.43 \\
              & \checkmark & &  &\checkmark & 75.59 & 5.73 & 1.84\\
             
             \bottomrule
        \end{tabular}
        \caption{Ablations on the high-resolution path and semantic guidance design on RCD. \textit{1/2}, \textit{1/4}, and \textit{1/8} denote the resolution of the high-resolution path relative to that of the original image, respectively. \textit{Single} indicates single-resolution guidance within the HrSeg block, while \textit{multi} indicates multi-resolution guidance.}
        \label{tab:table2}
    \end{table}
\end{center}

\subsubsection{Semantic guidance}
\label{subsubsec:semanticguidance}

As discussed in Section \ref{subsec:Semanticguidance}, we design two distinct schemes for extracting semantic information. One approach involves multi-resolution (see Figure \ref{fig:fig2} (b)) guidance within the HrSeg block, which is repeated three times. The other approach entails single-resolution (see Figure \ref{fig:fig2} (a)) guidance within the block but with the use of different resolution guidance paths across the three HrSeg blocks. Table \ref{tab:table2} displays the results of both semantic guidance methods. When compared to the single-path model, both of the semantic guidance schemes prove to be superior. At a resolution of 1/4 of the original image, the high-resolution path achieves 74.03\% mIoU. Furthermore, with a simple summation, both of the semantic guidance approaches yield improvements of 2.72\% and 1.56\%, respectively. This observation suggests that semantic guidance has a notable complementary effect on the features extracted through the high-resolution path. For the two different guidance manners, the computational cost of single-resolution guidance within the block is 40\% that of multi-resolution, and the parameter remains in a small range relative to the previous high-resolution model, HRNet \citep{wang_deep_2020}. Here, we adopt single-resolution semantic guidance within the block as the default.

\begin{center}
    \begin{figure}
        \centering
        \includegraphics[scale=0.5]{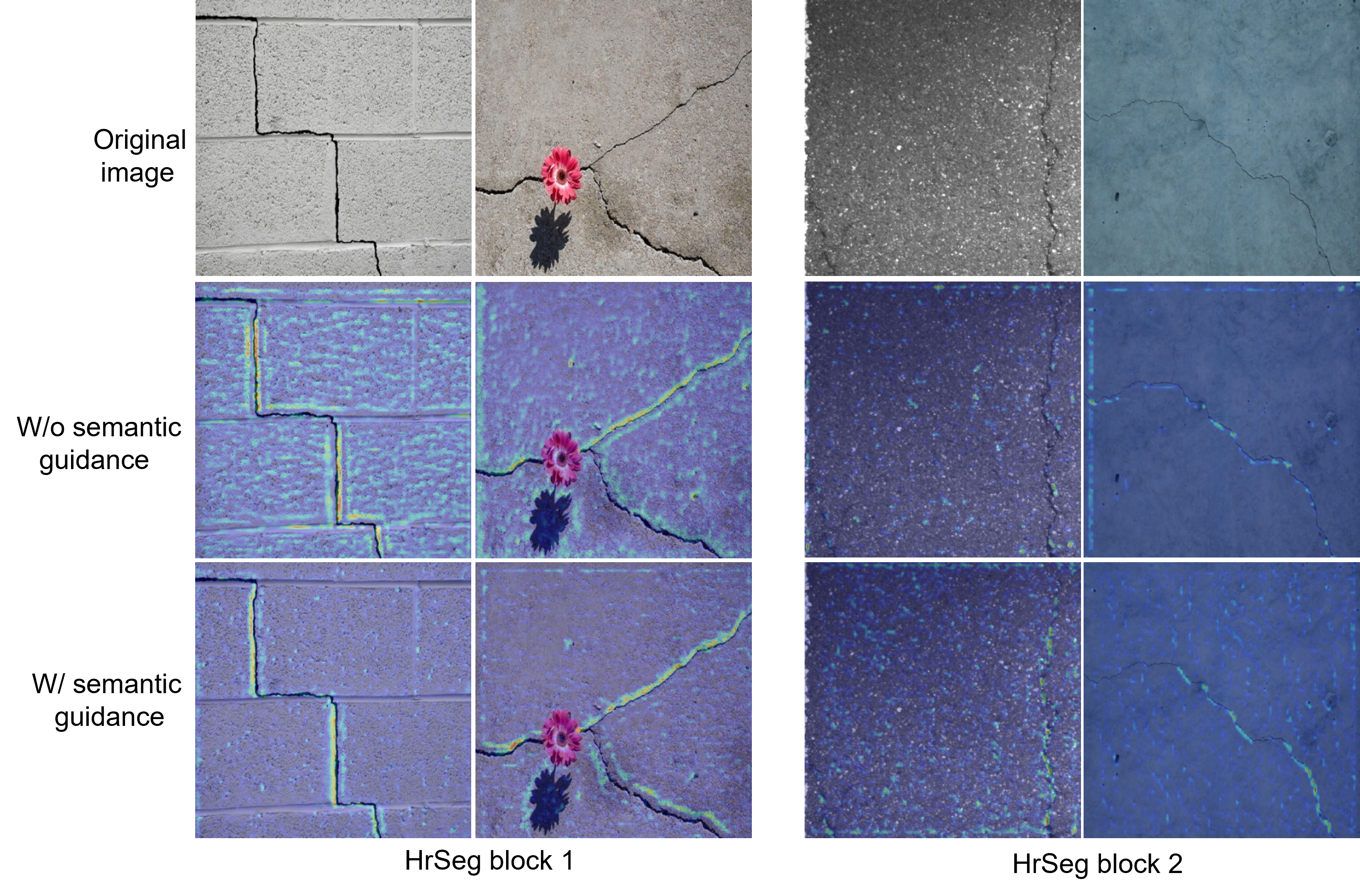}
        \caption{Examples showing visual explanations for the different stages of the HrSegNet. The first row depicts the original image, while the second and third rows illustrate the superimposed images without and with semantic guidance, respectively. The first and last two columns are obtained from HrSeg block 1 and 2.}
        \label{fig:fig5}
    \end{figure}
\end{center}

To better investigate the impact of semantic guidance on crack segmentation, we visualize the activation maps using Seg-Grad-CAM \citep{vinogradova_towards_2020}. The results are displayed in Figure \ref{fig:fig5}. The first and last two columns represent the two stages of HrSeg block 1 and 2. The first row shows the original image, while the second and third rows depict the Class Activation Map (CAM) visualizations without and with semantic guidance, respectively. It is clear that when semantic guidance is introduced, the HrSegNet can pay more attention to crack objects. In contrast, without semantic guidance, the model disperses its attention across the background (see first two columns in Figure \ref{fig:fig5}). Additionally, at different stages of the model, as it becomes deeper (HrSeg block 2 in Figure \ref{fig:fig5}), we observe that the model focuses more on small cracks when using semantic guidance, whereas, without semantic guidance, the model even struggles to detect them.

\subsubsection{Feature fusion}
\label{subsubsec:featurefusion}

\begin{center}
    \begin{table}[t]
        \centering
        \begin{tabular}{c| c| c c| c c |c c |c |c |c  }
            \toprule
            \multirow{2}{*}{HR} & \multirow{2}{*}{SG} & \multicolumn{2}{c|}{Fusion} & \multicolumn{2}{c|}{Seg head} & \multicolumn{2}{c|}{DS} & \multirow{2}{*}{OHEM} & \multirow{2}{*}{mIoU(\%)} & \multirow{2}{*}{GFLOPs} \\
             \cline{3-8}
             & & $\otimes$ & $\oplus$ & \textit{single} & \textit{double} & \textit{h1} & \textit{h2} &  & & \\
             \hline
             \checkmark & &  &  &  &  &  &  &  & 74.03 & 1.28\\
             \hline
              \checkmark & \checkmark & \checkmark  &  &  &  &  &  &  & 75.65 & 2.31 \\
              \checkmark & \checkmark &   &\checkmark  &  &  &  &  &  & 76.75 & 2.31 \\
              \hline
              \checkmark & \checkmark &   &\checkmark  & \checkmark &  &  &  &  & 76.75&2.31 \\
              \checkmark & \checkmark &   &\checkmark  &  & \checkmark  &  &  &  & 77.52& 2.49\\
              \hline
              \checkmark & \checkmark &   &\checkmark  &  & \checkmark  & \checkmark &  &  & 78.24& 2.49 \\
              \checkmark & \checkmark &   &\checkmark  &  & \checkmark  &  & \checkmark  &  &78.36 & 2.49\\
              \checkmark & \checkmark &   &\checkmark  &  & \checkmark  & \checkmark &  \checkmark&  &79.21 & 2.49\\
              \hline
              \checkmark & \checkmark &   &\checkmark  &  & \checkmark  & \checkmark &  \checkmark& \checkmark & 79.70 & 2.49 \\
             
             \bottomrule
        \end{tabular}
        \caption{Ablation study on the effectiveness of each component in HrSegNet. HR and SG represent the high-resolution path and semantic guidance, respectively. The segmentation head is evaluated in single-step and double-step modes. DS represents deep supervision, and \textit{h1} and \textit{h2} represent two different supervision locations in the HrSeg Block 1 and 2, respectively. OHEM is the online hard example mining.}
        \label{tab:table3}
    \end{table}
\end{center}

The fusion of features at different levels significantly impacts the result of semantic segmentation. For instance, when using vanilla semantic guidance, combining semantic and detailed information through summation improved the mIoU by 2.72\% (see Table \ref{tab:table3}). There are two mainstream methods for feature fusion: one is to fuse features of different positions during the model processing, such as skip connections used by UNet; the other is to fuse features before they enter last the segmentation head, such as PPM and ASPP. However, the latter is too heavy for real-time detection, so the fusion methods used in this paper are all carried out during the model processing.

BiSeNetV2 and DDRNet use a bilateral fusion strategy to merge high and low-level information to improve the feature extraction ability, but this structure leads to information redundancy. We use two simple yet practical fusion methods to reduce computational complexity and maximize semantic information guidance: element-wise multiplication and element-wise summation. Let $X_h$ and $X_s$ denote the high-level path and semantic-guided feature maps, respectively. These two fusion manners can be represented as follows:

\begin{equation}
    X_h = X_h \otimes Sigmoid(up(X_s))
\end{equation}

\begin{equation}
   X_h=  X_h \oplus  ReLU(up(X_s))
\end{equation}
\noindent $\otimes$ and $\oplus$ represent element-wise multiplication and element-wise summation, respectively. $up$ denotes up-sampling. We use different activation functions: sigmoid for element-wise multiplication and ReLU for element-wise summation.

The comparison of the results obtained from the two fusion strategies is presented in Table \ref{tab:table3}. Since both methods are point-wise operations, they have the same computational cost. The summation method outperforms the multiplication by 1.1\% in terms of mIoU, indicating its ability to provide better guidance for high-resolution details.

\subsubsection{Segmentation head}
\label{subsubsec:segmentationhead}

Our research proposes two forms of segmentation heads, single-step, and double-step, to generate high-resolution segmentation predictions through different up-sampling strategies. Specifically, the single-step segmentation head employs a single up-sampling operation to convert low-resolution feature maps to the same resolution as the input image. In contrast, the double-step segmentation head progressively up-samples the feature maps to the target resolution through two up-sampling operations.

In our experiments, we compared the performance of these two forms of segmentation heads, as shown in Table \ref{tab:table3}. The results demonstrate that, while the computational cost is almost the same for both forms, the double-step segmentation head outperforms the single-step segmentation head by 0.77\% in terms of mIoU. This suggests that the double-step up-sampling operation can better capture fine details. 

\subsubsection{Deep supervision}

\begin{figure}[t]
    \centering
    \includegraphics[scale=0.4]{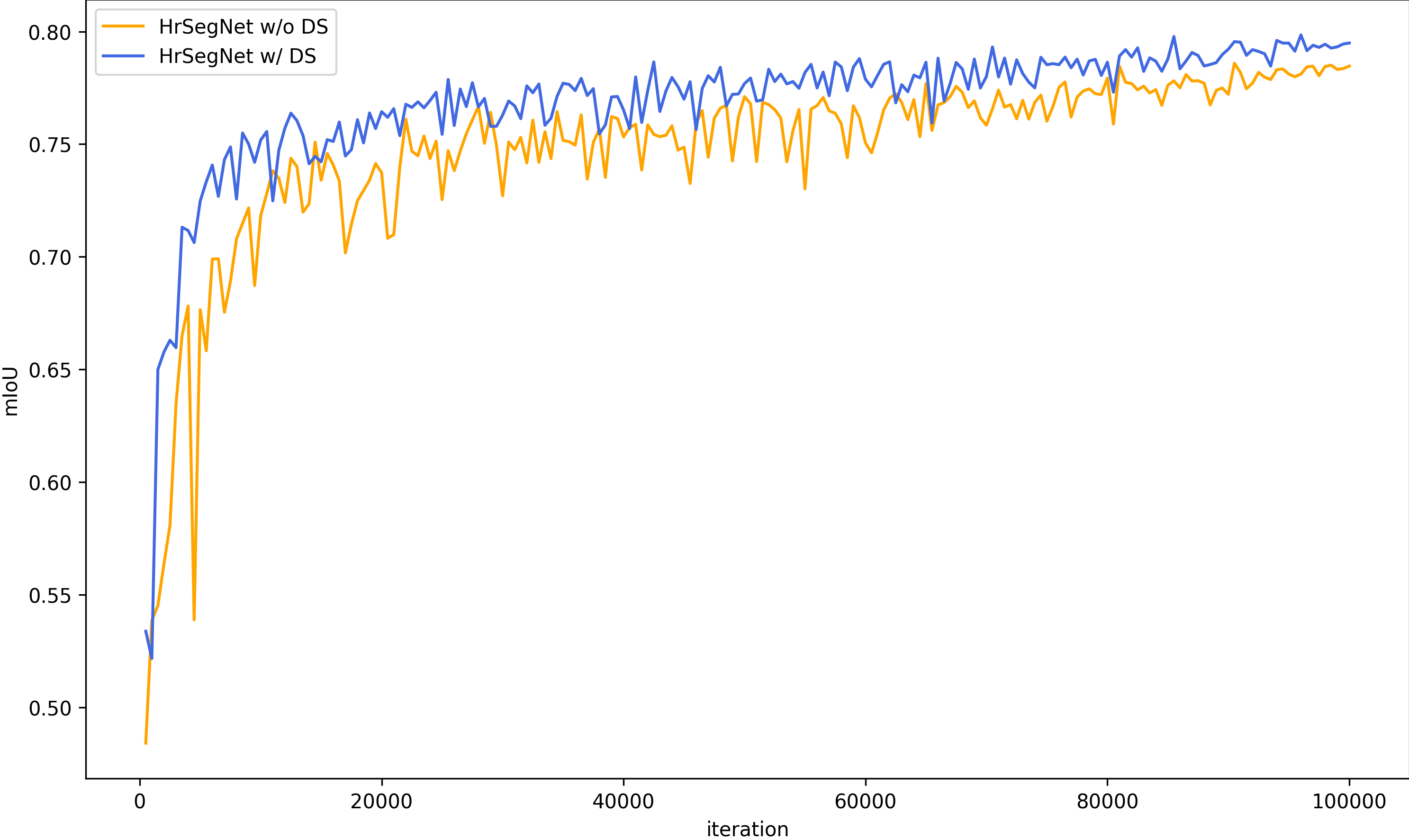}
    \caption{The mIoU curve of the training process for HrSegNet, with and without deep supervision.} 
    \label{fig:fig6}
\end{figure}

Deep supervision is only inserted into the high-resolution path during training and ignored during inference, so the additional heads do not affect inference efficiency. As shown in Table \ref{tab:table3}, we explore different positions for deep supervision. It is apparent that incorporating deep supervision results in an enhancement in segmentation accuracy without incurring any reduction in inference speed. Specifically, including deep supervision in HrSeg blocks 1 and 2 simultaneously yielded a 1.69\% mIoU increase. We conducted additional research on the convergence behavior of HrSegNet while utilizing deep supervision, illustrated in Figure \ref{fig:fig6}. The results indicate that incorporating deep supervision leads to a more rapid and stable convergence process, thereby substantially reducing the overall training time required.

\subsection{Scalability}

In this section, we delve into the proposed structure's scalability. As we introduce in Section \ref{subsec:Overallarchitecture}, our model is designed to be very flexible for real-time applications. We can easily generalize it to larger or smaller models by adjusting the capacity of the high-resolution path. Table \ref{tab:table5} showcases the quantization results for the segmentation models with varying computational complexities. The fastest model we proposed, HrSegNet-B16, achieves an inference speed of 182 FPS and 78.43\% mIoU on the benchmark CrackSeg9k, with a computational complexity of 0.66 GFLOPs. The model with the highest accuracy, HrSegNet-B48, achieves 80.32\% mIoU at 140.3 FPS, with a computational complexity of 5.60 GFLOPs.


\subsection{Stability and robustness}

\begin{center}
    \begin{figure}[t]
        \centering
        \includegraphics[scale=0.5]{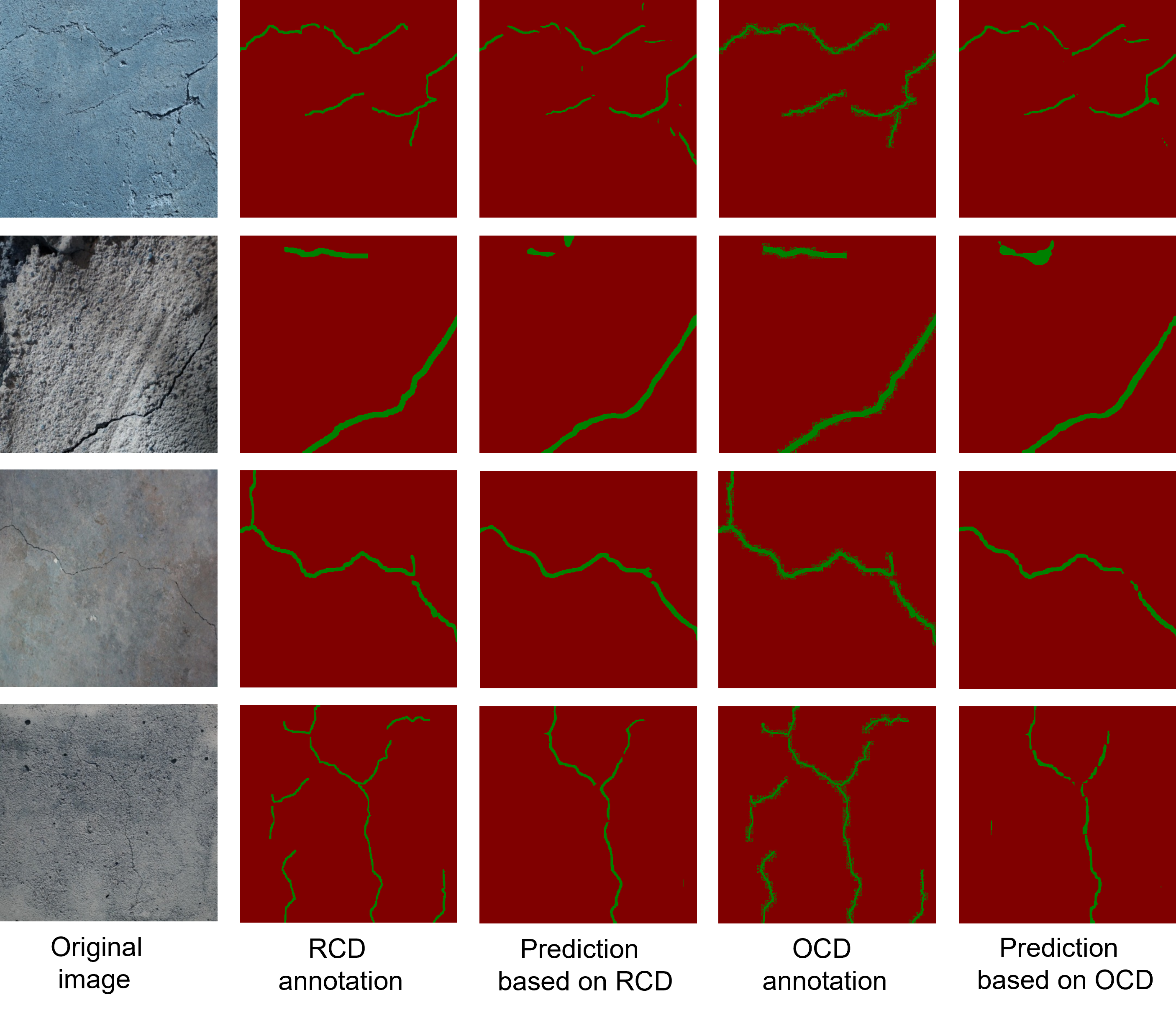}
        \caption{Visualization examples on the RCD and OCD test set. The first column depicts the original image, while the second and third columns represent pseudo-colored labels from two datasets. The fourth and fifth columns display the predictions of HrSegNet-B32. It is worth noting that both datasets' labels contain annotation errors in the last row of images, but our model provides accurate predictions.}
        \label{fig:figure7}
    \end{figure}
\end{center}

We assess the stability and robustness of our model in this section. As described in Section \ref{subsec:datasets}, OCD and RCD share seven sub-datasets, but the annotations in OCD are more noisy and challenging. We train the HrSegNet-B32 on both datasets, and the quantization results are shown in Table \ref{tab:table4}. Our model achieves the mIoU score that is on par with the performance obtained on the precisely annotated dataset RCD when applied to the dataset OCD which has more noisy annotations. Qualitative results of the model on the test set are shown in Figure \ref{fig:figure7}. The first column displays the input images, while the second and fourth columns show the pseudo-color annotations of the same sample on RCD and OCD, respectively. It can be clearly seen that the annotations of OCD contain more noise, with hair-like boundary distortions around the cracks. This caused scenarios where the model's metric score is lower even for visually accurate predictions due to the distortions in the ground truth. The third and fifth columns display the model's prediction on the two datasets, respectively. Despite the presence of noise, our model exhibits consistent performance and produces stable results even in the presence of noisy conditions. In situations where errors occur in ground truth (see the last row of Figure \ref{fig:figure7}), both models accurately predict the location of actual crack pixels.

\begin{center}
    \begin{table}[t]
        \centering
        \begin{tabular}{c|c}
            \toprule
             Dataset & mIoU(\%)  \\
             \hline
             RCD & 79.70\\
             OCD & 79.23\\
            \bottomrule
        \end{tabular}
        \caption{The mIoU scores of HrSegNet-B32 on RCD and OCD.} 
        \label{tab:table4}
    \end{table}
\end{center}

\subsection{Comparisons with state-of-the-art}

Our objective is to attain a superior trade-off between accuracy and speed. Thus our emphasis lies in achieving high segmentation accuracy while maintaining real-time inference. In this section, we compare the results of our models with seven segmentation models on the RCD test set. We utilize the RCD training and validation sets to train all the models and evaluate their segmentation accuracy on the test set. Inference time measurements are conducted on NVIDIA GeForce RTX 2070 SUPER with TensorRT 8.6. For the sake of efficient and expedient comparisons, our training is conducted from scratch without any pre-training on other datasets.

\begin{center}
    \begin{table}[t]
        \centering
        \begin{tabular}{c|c |c |c |c}
        \toprule
             Model & mIoU(\%) & FPS & Params & GFLOPs \\
        \hline
             UNet & 76.71 & 41.5 &13.40 & 75.87\\ 
             PSPNet(ResNet18) & 78.10 & 59.5 & 21.07& 54.20\\ 
             BiSeNetV2 & 78.37& 77.1 &2.33 &4.93 \\ 
             STDCSeg(STDC1) & 78.33 & 82.3 & 8.28& 5.22\\ 
             DDRNet & 76.58& 122& 20.18& 11.11\\ 
             OCRNet(HRNet-W18) & 80.90& 39 & 12.11& 32.40\\ 
             DeeplabV3+(ResNet18) & 78.29& 60.6&12.38 &33.96 \\ 
             HrSegNet-B16 & 78.43& 182& 0.61& 0.66\\ 
             HrSegNet-B32 & 79.70& 156.6& 2.49&2.50 \\ 
             HrSegNet-B48 &80.32 & 140.3&5.43 &5.60 \\ 
        \bottomrule
        \end{tabular}
        \caption{Comparisons with state-of-the-art on RCD.}
        \label{tab:table5}
    \end{table}
\end{center}

Table \ref{tab:table5} compares our method and state-of-the-art. Our HrSegNet achieves excellent inference speed while maintaining competitive segmentation accuracy. Specifically, our most miniature model, HrSegNet-B16, achieves 78.43\% mIoU on the self-divided RCD test set at a speed of 140.3 FPS, outperforming PSPNet \citep{zhao2017pyramid}, BiSeNetV2 \citep{yu_bisenet_2020}, STDCSeg \citep{fan_rethinking_2021}, and DeeplabV3+ \citep{chen_rethinking_2017}with similar accuracy. Moreover, the computational complexity of HrSegNet-B16 is remarkably efficient, equivalent to only 13.4\% and 12.6\% of the state-of-the-art real-time semantic segmentation models, BiSeNetV2 and STDCSeg, respectively. HrSegNet-B16 only requires 0.66 GFLOPs of computational cost, making it very lightweight. The medium-sized model, HrSegNet-B32, achieves a performance improvement of 1.27\% compared to the smaller one. Although the parameters and computational complexity have increased fourfold, the model still maintains a very fast real-time segmentation speed at 156.6 FPS. We increase the channel capacity of the HrSeg block to 48, which is HrSegNet-B48, resulting in a segmentation accuracy improvement of 0.62\%. While the parameters and computational complexity doubled, it still meets real-time segmentation requirements and achieves 140.3 FPS.

\begin{center}
    \begin{figure}
        \centering
        \includegraphics[scale=0.3]{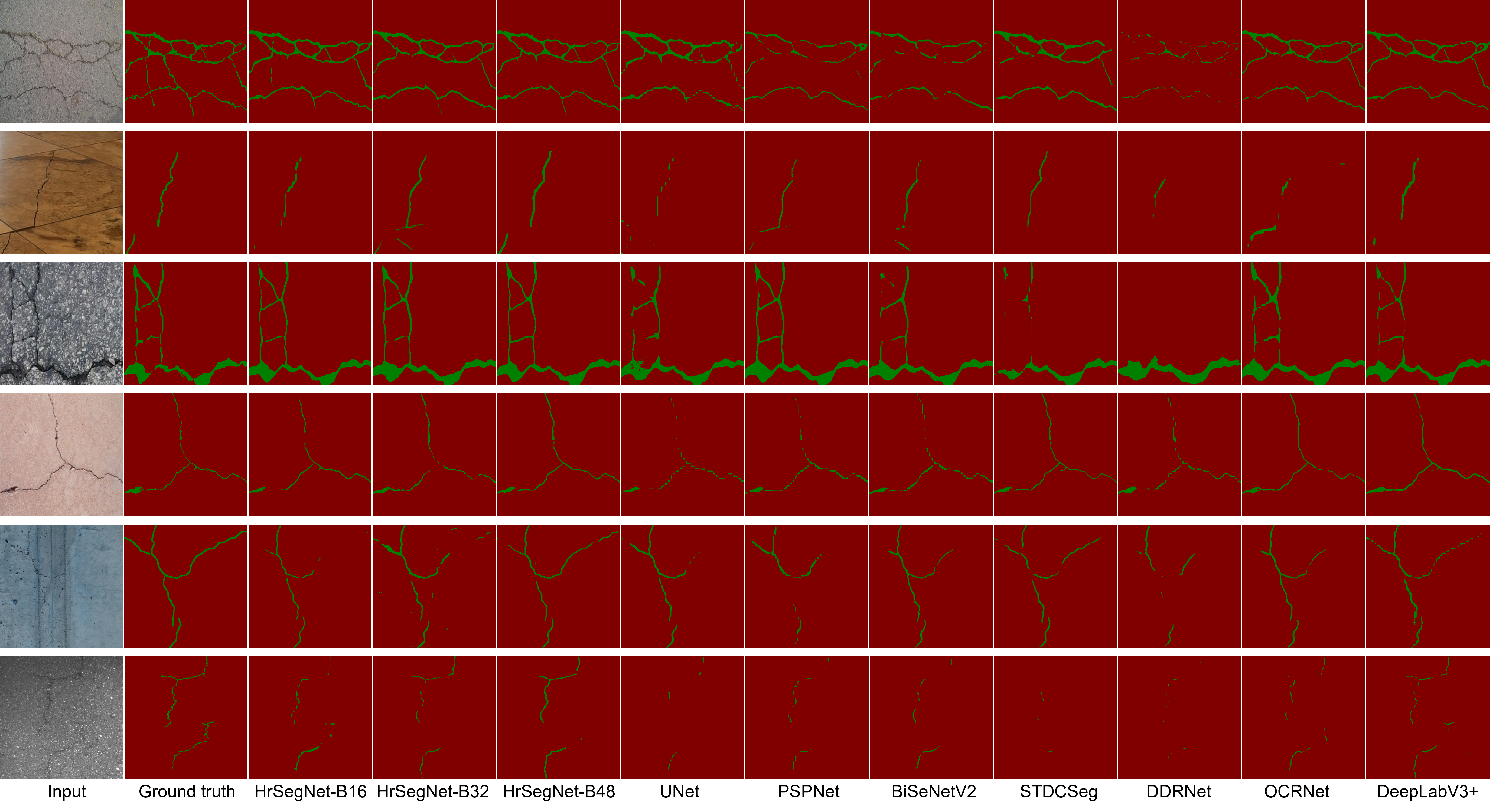}
        \caption{Visualized segmentation results on RCD test set.}
        \label{fig:fig8}
    \end{figure}
\end{center}

Comparative experiments reveal that despite its highest computational complexity of 75.87 GFLOPs, UNet \citep{ronneberger_u-net_2015} only attains a segmentation accuracy of 76.71\%, which is the lowest among all models. DDRNet \citep{pan_deep_2023}, similar to our structure, only achieves 76.58\% mIoU on the RCD test set. OCRNet \citep{vedaldi_object-contextual_2020}, which uses HRNet-W18 as the backbone, achieved the highest 80.90\% mIoU. However, as discussed in Section \ref{subsec:highper}, HRNet is very heavy and complex, and the inference speed is difficult to achieve real-time segmentation requirements. As CrackSeg9k does, we also test DeeplabV3+, but they use ResNet101 as the backbone, while we use ResNet18 because ResNet101 cannot meet the real-time requirements. In our test, DeeplabV3+ achieves 78.29\% mIoU at 60.6 FPS but still lags behind our HrSegNet. In order to emphasize the effectiveness of our method, we show some examples of the RCD test set in Figure \ref{fig:fig8}.

\section{Concluding remarks}
We observe that segmenting cracks requires a high-resolution representation and supplementary contextual information. We devise a novel architecture named HrSegNet, which efficiently and parallelly processes high-level and low-level information, thereby merging them. The HrSegNet exhibits high scalability, yielding state-of-the-art segmentation accuracy and significantly outperforming state-of-the-art models in terms of inference speed, as demonstrated on the CrackSeg9k dataset. Compared with popular segmentation models, we find that excessive design for crack segmentation is ostentatious and impractical. Our design, on the other hand, is intuitive, versatile, and remarkably effective. We hope this research will contribute to advancements in the field of crack segmentation.

\section*{CRediT authorship contribution statement}
\textbf{Yongshang Li:} Conceptualization, Methodology, Writing – original draft, Writing – review \& editing, Investigation, Validation. \textbf{Ronggui Ma:} Resources, Supervision, Funding acquisition, Writing – review \& editing, Project administration. \textbf{Han Liu:} Investigation, Writing – review \& editing. \textbf{Gaoli Cheng:} Resources, Supervision, Funding acquisition.

\section*{Funding}
This work was supported in part by the Key Research and Development Project of China under Grant 2021YFB1600104, in part by the the National Natural Science Foundation of China under Grant 52002031, and also in part by the Scientific Research Project of Department of Transport of Shaanxi Province under Grants 20-24K, 20-25X.

\section*{Declaration of Competing Interest}
The authors declare that they have no known competing financial interests or personal relationships that could have appeared to influence the work reported in this paper.

 \bibliographystyle{elsarticle-harv} 
 \bibliography{pp2}





\end{document}